%% file: bare_jrnl.tex
\documentclass[journal]{IEEEtran}
\usepackage{cite}

\ifCLASSINFOpdf
\else
\fi
\usepackage{amssymb}
\usepackage{amsmath}
\usepackage{multirow}
\usepackage{graphicx}
\usepackage{booktabs} 
\usepackage{tabularx}
\usepackage[normalem]{ulem}
\useunder{\uline}{\ul}{}
\usepackage{placeins}

\usepackage{hyperref}

\begin{document}
%
\title{Slot-MLLM: Object-Centric Visual Tokenization for Multimodal LLM}
%
%
%


\author{
Donghwan Chi,
Hyomin Kim,
Yoonjin Oh,
Yongjin Kim,
Donghoon Lee,
Daejin Jo,
Jongmin Kim,
Junyeob Baek,
Sungjin Ahn,
and Sungwoong Kim%
\thanks{
Donghwan Chi, Hyomin Kim, Yoonjin Oh, Yongjin Kim, Daejin Jo, and Sungwoong Kim are with the Department of Artificial Intelligence, Korea University, Seoul, Republic of Korea.
}%
\thanks{
Donghoon Lee and Jongmin Kim are with Kakao Corp.
}%
\thanks{
Junyeob Baek and Sungjin Ahn are with the School of Computing, KAIST, Daejeon, Republic of Korea.
}%
\thanks{
Corresponding authors: Sungjin Ahn and Sungwoong Kim
(e-mail: sungjin.ahn@kaist.ac.kr; swkim01@korea.ac.kr).
}}

\maketitle

\begin{abstract}
\input{sections/abstract}
\end{abstract}


%
\IEEEpeerreviewmaketitle

%
%
%
%

 

\section{Introduction}
\input{sections/introduction}

\section{Related Work}
\input{sections/related_work}

\section{Method}

\input{sections/method}

\section{Experiments}
\input{sections/experiments}

\section{Conclusion}

\input{sections/conclusion}

\section*{Acknowledgment}
\input{sections/acknowledgments}

\ifCLASSOPTIONcaptionsoff
  \newpage
\fi



%
\bibliographystyle{IEEEtran}
\bibliography{references}



%








\end{document}

%% file: sections/abstract.tex
Recent generative vision-language MLLMs adopt discrete visual tokens to model images autoregressively alongside text, but existing tokenizers are largely patch-based or dominated by global semantics, limiting object-level understanding and precise controllable generation. To address this, we introduce \textbf{SlotTok}, an object-centric tokenizer that makes slot attention practical for generative MLLMs by encoding images into a compact set of slot-based visual tokens while preserving both semantic structure and reconstruction quality. SlotTok is trained with image reconstruction, image–text alignment, and attention guidance losses, and paired with a diffusion-based decoder to enable faithful image synthesis from slot tokens. Building on SlotTok, we develop \textbf{Slot-MLLM}, which supports unified autoregressive modeling for understanding and generation. Extensive experiments show that Slot-MLLM consistently improves object-centric understanding and fine-grained controllable generation, including localized image editing, compared to MLLMs based on prior visual tokenizers. To our knowledge, this work provides the first large-scale empirical validation that object-centric slot attention can be effectively integrated into generative MLLMs on real-world images.

%% file: sections/introduction.tex
\begin{figure*}[t!]
    \centering
    \includegraphics[width=0.95\linewidth]{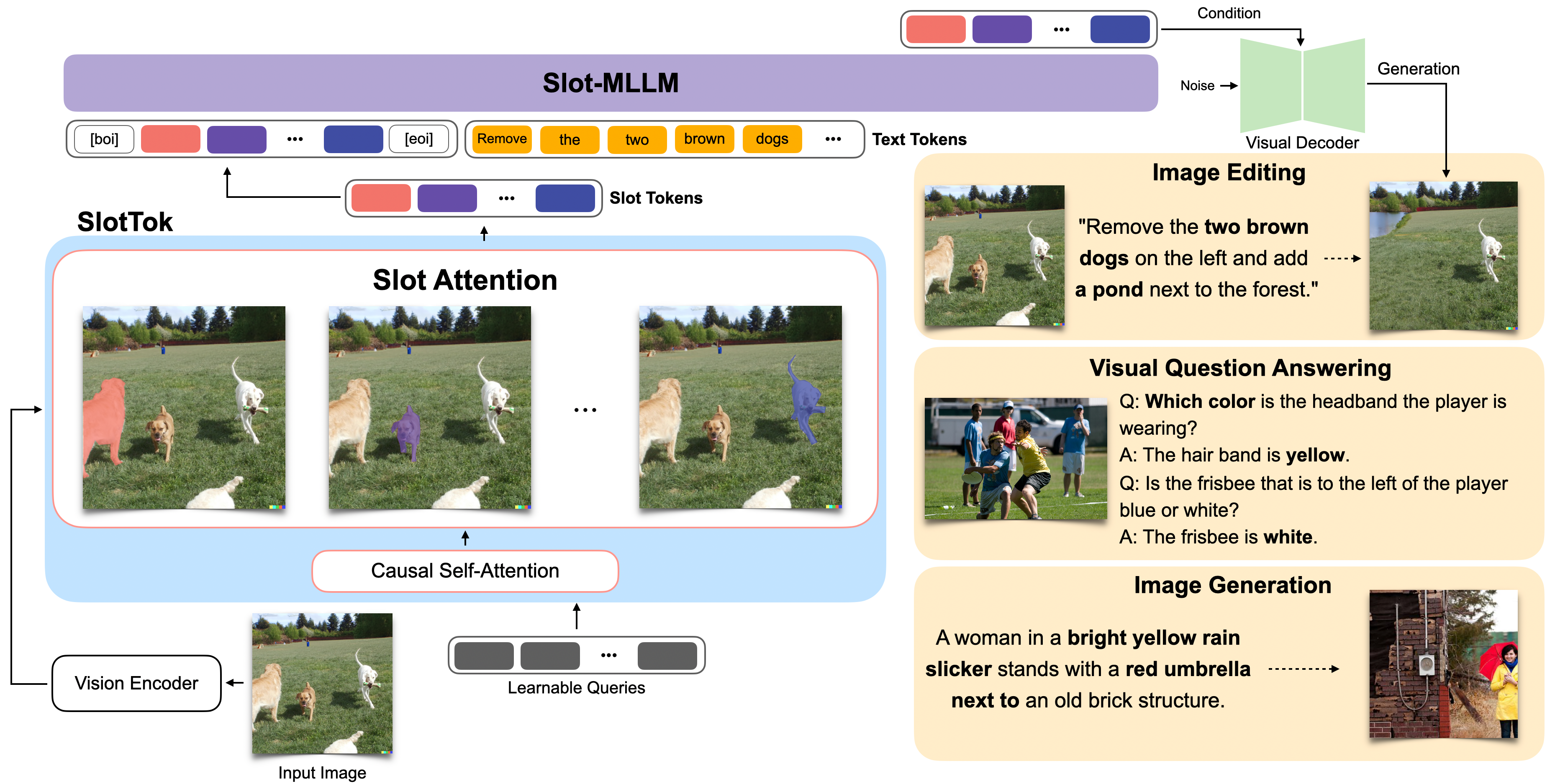}
    \caption{Overview of Slot-MLLM. Slot-MLLM employs SlotTok with slot attention to encode images into discrete, object-centric visual tokens. These tokens are treated as a second language within the model, enabling unified autoregressive modeling across multimodal tasks such as image editing (IT2I), visual question answering (I2T), and image generation (T2I).\vspace{-1.0em}}
    \label{fig:main-figure}
\end{figure*}


Recent vision-language multimodal large language models (MLLMs) adopt \emph{discrete} image tokens to enable autoregressive modeling of images alongside text \cite{ge2023making, jin2023unified, wu2024vila, pan2024auto, xie2025musevl}. These models are \emph{unified} in that a single MLLM jointly supports both multimodal understanding and image generation within the same autoregressive next-token prediction framework. 
Although this discrete token paradigm is attractive for unified generation, existing visual tokenizers often struggle to preserve structural compositionality, which is critical for object-level reasoning and controllable multimodal generation. Although recent unified tokenizers like UniTok \cite{ma2025unitok} and TokenFlow \cite{qu2025tokenflow} attempt to bridge the gap between high-level semantics and pixel reconstruction, they primarily rely on global semantic supervision within a patch-based framework. Consequently, these approaches remain constrained by fixed-grid representations that result in excessively long visual sequences. Furthermore, in the case of TokenFlow, the architecture depends on decoupled models for understanding and generation, which limits its potential as a fully integrated MLLM. 
Separately, other efforts to inject locality have relied on patch-level tokens with heuristic clustering or merging \cite{jin2023unified,wu2024towards}. These methods often involve ad-hoc grouping processes on top of fixed patch features, rather than discovering entities through a principled, unsupervised mechanism.

Slot attention \cite{locatello2020object} provides a principled inductive bias for object-centric representation learning by decomposing a scene into a fixed set of slot tokens.
However, making it practical as a visual tokenizer for generative MLLMs is challenging because it must simultaneously satisfy several demanding requirements: (i) it must support high-quality image generation, ensuring that slot tokens can be faithfully decoded into realistic images; (ii) it must preserve strong object-level structure and also align well with language semantics so that visual tokens can be effectively integrated into MLLMs; and (iii) it must remain stable and scalable when trained on large-scale, diverse natural image–text datasets. While slot attention is particularly appealing in this setting due to its ability to compress visual content into a small set of semantically meaningful tokens, meeting all these requirements within a unified tokenizer has remained an open challenge.

To this end, we introduce \textbf{SlotTok}, an object-centric visual tokenizer designed to enable unified multimodal modeling and token-efficient image representation. Using slot attention, SlotTok explicitly preserves object-level structure while compressing visual content into a small set of discrete tokens, allowing images to be represented with significantly fewer tokens while preserving semantic and structural fidelity for both reasoning and generation. To make these object-centric tokens practical for integration into MLLMs, we incorporate dedicated architectural and training choices, including attention guidance loss to stabilize slot-object alignment and image--text alignment loss to ground slots in language semantics. Together, these components ensure that SlotTok maintains strong objectness while remaining compatible with autoregressive multimodal modeling. 


Building on SlotTok, we develop \textbf{Slot-MLLM}, which operates under a unified next-token prediction paradigm for both modalities. Extensive experiments show that Slot-MLLM maintains competitive general performance while significantly improving fine-grained structural control. This unified design highlights that object-centric tokenization is not just a representational choice, but a key enabler for efficient and controllable multimodal generation in large-scale MLLMs.

In summary, our main contributions are as follows.
(1) We propose \textbf{SlotTok}, an object-centric visual tokenizer designed to enable a unified and efficient visual interface for MLLMs.
(2) We introduce a practical training strategy for SlotTok that allows slot-based tokenization to be learned without degrading pretrained diffusion priors.
(3) We present \textbf{Slot-MLLM}, providing the first large-scale empirical validation that slot attention can be effectively integrated into LLMs for diverse real-world images, demonstrating consistent improvements over existing baselines in multimodal understanding and generation.

\begin{figure*}[t]
    \centering
    \includegraphics[width=0.95\textwidth]{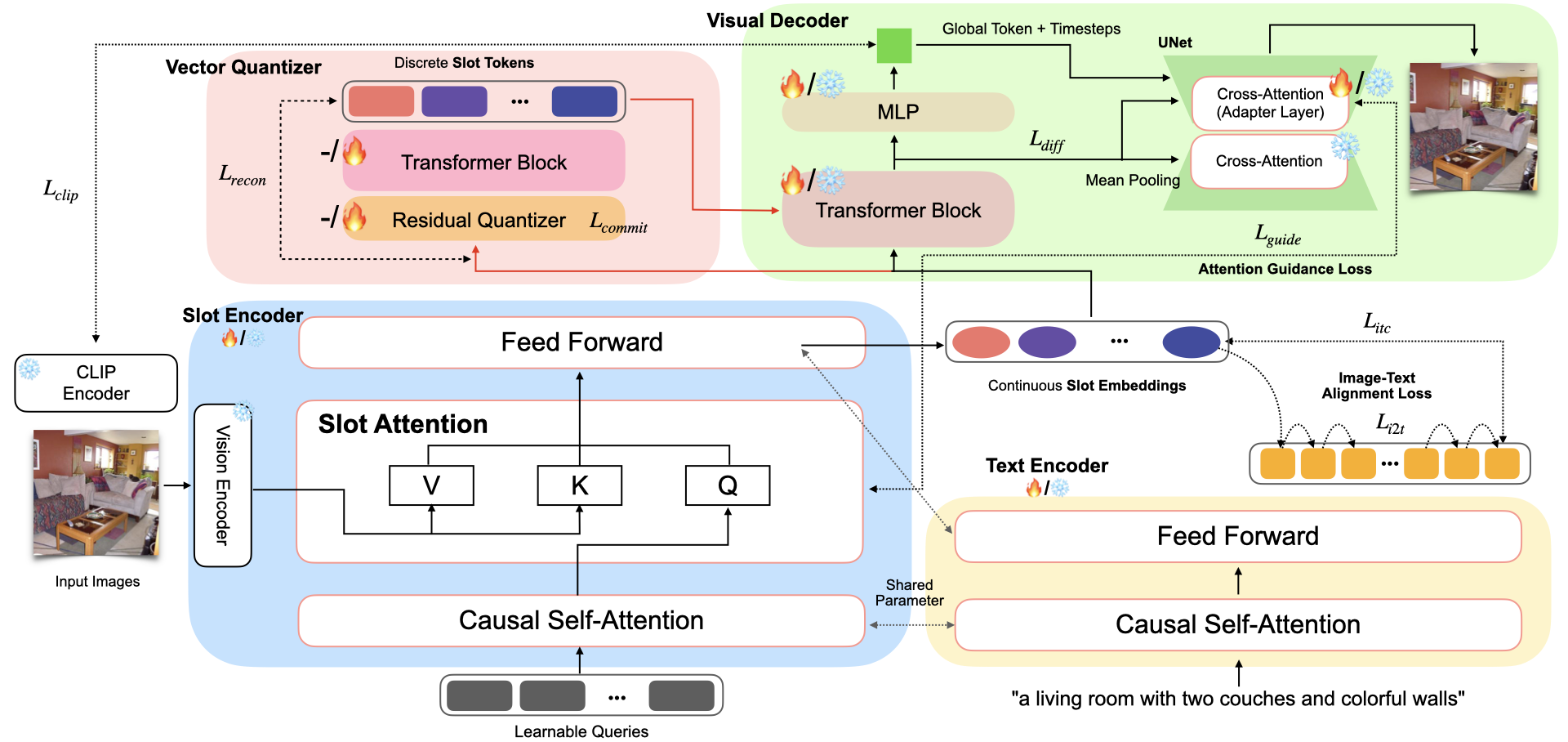}
    \caption{Our tokenizer comprises three main modules: Slot Encoder, Vector Quantizer, and Visual Decoder. Training occurs in two stages: modules trained in each stage are indicated, and modules exclusive to the second stage—marked in red—include the Vector Quantizer for discrete tokenization.\vspace{-1.0em}}
    \label{fig:training-overview}
\end{figure*}

%% file: sections/related_work.tex

\textbf{Multimodal Large Language Models.} 
Recent multimodal large language models (MLLMs) extend LLMs to visual inputs by connecting pretrained vision encoders with language backbones. Representative understanding-oriented models, such as BLIP-2~\cite{li2023blip} and LLaVA~\cite{liu2023visual}, project visual features into the language embedding space and focus mainly on visual instruction following and multimodal reasoning. In contrast, generative MLLMs aim to unify understanding and image generation by modeling images as discrete visual tokens alongside text. Models such as SEED-LLaMA~\cite{ge2023making}, LaViT~\cite{jin2023unified}, DreamLLM~\cite{dong2023dreamllm}, Emu~\cite{sun2024emu}, LWM~\cite{liu2024world} and VILA-U~\cite{wu2024vila} demonstrate this paradigm, while recent systems such as Janus-Pro~\cite{chen2025janus} and BAGEL~\cite{deng2025emerging} further improve generation quality through larger-scale training, stronger backbones, and additional alignment strategies. However, these advances do not directly address how visual tokens should be structured for object-level reasoning and controllable generation.


\textbf{Visual Tokenization for Unified Understanding and Generation.}
Visual tokenization determines what visual information is exposed to the autoregressive model. Early discrete tokenizers, including VQ-VAE~\cite{oord2018neural}, VQGAN-style tokenizers, and residual quantization methods~\cite{lee2022autoregressive}, are mainly optimized for image reconstruction, but their patch-level codes often lack explicit semantic organization. Recent unified tokenizers improve semantic alignment between visual tokens and language: SEED~\cite{ge2023making} introduces causal visual tokens with a diffusion decoder, and UniTok~\cite{ma2025unitok} and TokenFlow~\cite{qu2025tokenflow} further improve compatibility with MLLMs. Nevertheless, these methods still largely rely on fixed-grid or globally supervised representations. Local semantic approaches such as SeTok~\cite{wu2024towards} introduce clustering or token merging, but depend on explicit grouping heuristics. In contrast, SlotTok uses slot attention as a principled inductive bias to learn compact object-aware visual tokens for both understanding and generation.

\textbf{Slot Attention.} Slot attention \cite{locatello2020object} effectively generates unsupervised object-centric image representations. Previous works like LSD \cite{jiang2023object} have successfully combined slot attention with generative models for reconstruction tasks. Although previous works have explored slot attention, their performances have not been thoroughly demonstrated at large scale, particularly in integration with LLMs and in-the-wild images. MorphTokens~\cite{pan2024auto} also attempted to include slot attention, but clear improvements through slot attention have not been well established, and moreover, visual tokens for image generation are separated from those for image understanding.

%% file: sections/method.tex
The primary goal of this work is to make slot attention practical for generative MLLMs by turning it into an object-centric visual tokenizer that supports multimodal understanding and image generation. To this end, we propose \textbf{SlotTok}, which encodes an image into a compact set of object-centric slot representations. Building on SlotTok, we develop \textbf{Slot-MLLM}, which integrates these tokens into a single next-token prediction framework. The overall structure of Slot-MLLM with SlotTok is depicted in Fig.~\ref{fig:main-figure}.

\subsection{SlotTok: Object-Centric Visual Tokenizer}

We propose \textbf{SlotTok}, an object-centric visual tokenizer that enables \emph{unified} autoregressive multimodal modeling while remaining \emph{token-efficient}. The overall architecture of SlotTok is illustrated in Fig.~\ref{fig:training-overview}. SlotTok comprises (i) Slot Encoder for object-centric slot embeddings, (ii) Vector Quantizer for discrete tokenization, and (iii) Visual Decoder for image synthesis.

SlotTok is trained in two stages. In the first stage, Slot Encoder is trained to produce \emph{continuous} slot embeddings that capture object-level details while being aligned with language semantics. In the second stage, these continuous embeddings are quantized into \emph{discrete} slot tokens, which facilitate a next-token prediction \cite{radford2018improving} for both understanding and generation. Importantly, SlotTok is trained independently from the LLM, allowing flexible integration of various pre-trained LLM backbones. 

\subsubsection{Architecture}

\textbf{Slot Encoder.} Given an input image $\mathbf{x}\in \mathbb{R}^{H \times W \times C}$, Slot Encoder maps the image into a set of $N$ \emph{continuous slot embeddings} $S_{emb} \in \mathbb{R}^{N \times D}$, where each slot is intended to represent a distinct object or region in the scene. To this end, Slot Encoder is built on a query-based transformer architecture. The input image $\mathbf{x}$ is first processed by a vision encoder $f^{\text{Enc}}$, yielding a set of visual features $E = f^{\text{Enc}}(\mathbf{x}) \in \mathbb{R}^{M \times D_{\text{input}}}$. Conditioned on these features, Slot Encoder employs slot attention to aggregate visual information into $N$ coherent slots through competitive attention.

Specifically, learnable slot queries $S \in \mathbb{R}^{N \times D}$ are initialized from a Gaussian distribution. Each slot attends to the visual features $E$ as keys and values, producing attention weights
\begin{equation}
A^{SA}=\text{softmax}_N\left(\frac{q(S) \cdot k(E)^T}{\sqrt{D}}\right),
\end{equation}
which are normalized across slots to assign each visual feature to a single slot:
Slot representations are progressively refined in a \emph{layer-wise} manner. After all layers, the final outputs are the slot embeddings $S_{emb}=\{s_1,s_2,\dots,s_N\}$.

To facilitate seamless integration with autoregressive MLLMs, Slot Encoder additionally employs a causal self-attention layer following SEED \cite{ge2023making}. This design aligns slot embeddings with the token-level causality assumed by language models, ensuring that visual representations can be generated under the same next-token prediction paradigm as text.

\textbf{Vector Quantizer.} Continuous slot embeddings $S_{emb}$ produced by the Slot Encoder are discretized into \emph{slot tokens}. We adopt a Residual Vector Quantizer (RVQ) \cite{lee2022autoregressive}, together with a lightweight Transformer block that reconstructs slot embeddings from the discrete codes. This design improves reconstruction performance compared to standard vector quantization, which we find crucial for preserving image quality under a compact token budget. Detailed implementation is provided in the supplemental material.


\textbf{Visual Decoder.} Our Visual Decoder consists of a transformer decoder block, an MLP head, and an unCLIP Stable Diffusion (unCLIP-SD) model \cite{ramesh2022hierarchical}. Given slot embeddings $S_{emb}$ from Slot Encoder, we first obtain refined embeddings $\hat{S}=f^{Dec}(S_{emb})$. We then inject the slot information into unCLIP-SD through three conditioning pathways. First, a global conditioning vector $\hat{S}_{global}$ is produced by an MLP head from $\hat{S}$ and added to the timestep embedding to provide global semantic guidance. Second, for the original UNet \cite{ronneberger2015u} cross-attention layers, we construct a slot context embedding $\hat{S}_{\text{ctx}}=\text{MeanPool}(\hat{S})$ and provide $\hat{S}_{\text{ctx}}$ as the condition while keeping these pretrained cross-attention layers \emph{frozen} to preserve diffusion prior. Third, after each frozen cross-attention layer, we insert an additional \emph{adapter cross-attention} layer that is \emph{trainable} and receives the full set of slot embeddings $\hat{S}$.

\subsubsection{SlotTok Training} \label{method_stage1}

As illustrated in Fig.~\ref{fig:training-overview}, the first stage of tokenizer training involves learning Slot Encoder, Transformer Block, MLP, and trainable adapter cross-attention layers within the UNet of unCLIP-SD, while keeping the original UNet cross-attention layers frozen. Slot Encoder is initialized with weights from BLIP-2 \cite{li2023blip}. The first training stage aims to produce continuous slot embeddings $S_{emb}$ and is guided by three primary objectives: image reconstruction, image--text alignment, and attention guidance loss.

\textbf{Image Reconstruction Loss.} The global token $\hat{S}_{global}$ is trained to be aligned with the CLIP \cite{radford2021learning} image embedding $I$ using Mean Squared Error (MSE) loss.
\begin{equation}
    \mathcal{L}_{clip}=||\hat S_{global}-I||^2.
\end{equation}


Using three conditioning pathways---the global timestep condition $\hat{S}_{global}$, the slot context condition $\hat{S}_{\text{ctx}}$ for the frozen UNet cross-attention layer, and the full slot embeddings $\hat{S}$ for the trainable adapter cross-attention layer---we train the denoising process of unCLIP-SD in the Visual Decoder. Following standard diffusion training \cite{ho2020denoising, rombach2022high}, the UNet $g^{\text{UNet}}$ is trained to predict the noise $\hat{\epsilon}_t$ added to the latent representation $z_t$ at timestep $t$ as $\hat{\epsilon}_t = g^{\text{UNet}}(z_t, t+\hat{S}_{global}; \hat{S}_{\text{ctx}}, \hat{S})$, with 
\begin{equation}
    \mathcal{L}_{diff} = \lVert \hat{\epsilon}_t - \epsilon_t \rVert^2.
\end{equation}


\textbf{Image--Text Alignment Loss.} Despite effective image reconstruction, slot embeddings should be compatible with textual tokens for efficient training with LLMs. To ensure alignment between visual and textual modalities, we introduce two additional objectives: Image--Text Contrastive Loss ($\mathcal{L}_{itc}$) and Image-Grounded Text Generation Loss ($\mathcal{L}_{i2t}$). Inspired by BLIP-2 \cite{li2023blip}, we share parameters between the causal self-attention layer and feed forward layer in Slot Encoder and the text encoder, facilitating a better textual alignment of slot embeddings.

Specifically, unlike BLIP-2 and SEED \cite{ge2023making}, we maintain causal masking within self-attention when encoding text, preserving token-level causality. The resulting final slot embedding $s_N$ and final text embedding $\ell$ are used for contrastive learning \cite{oord2018representation} to further enhance the alignment and causal structure of slot embeddings. 
\begin{equation}
\begin{aligned}
\mathcal{L}_{itc}
=&-\log\frac{\exp (s_N^{(i)\top} \ell^{(i)}/\tau)}
{\sum_{j=1}^{B}\exp(s_N^{(i)\top} \ell^{(j)}/\tau)} \\
&-\log\frac{\exp (\ell^{(i)\top} s_N^{(i)}/\tau)}
{\sum_{j=1}^{B}\exp(\ell^{(i)\top} s_N^{(j)}/\tau)},
\end{aligned}
\end{equation}
where $B$ and $\tau$ denote batch size and learnable temperature parameter, respectively.
Image-Grounded Text Generation Loss follows the autoregressive text prediction approach of BLIP-2 by predicting text tokens conditioned on slot tokens.
\begin{equation}
    \mathcal{L}_{i2t}=-\sum_i \log P(y_i|s_{1},\dots,s_{N},y_{1},\dots,y_{i-1}),
\end{equation}
where $y_i$ denotes the $i$-th text token. 


\textbf{Attention Guidance Loss.} To encourage object-centric slot formation and ensure that the diffusion decoder consumes slots in an object-aligned manner, we introduce an \emph{attention guidance loss} that aligns the adapter cross-attention of the UNet with the slot attention patterns produced by Slot Encoder. This idea is inspired by SlotAdapt \cite{akan2025slot}, which uses adapter cross-attention masks as a self-supervisory signal to guide slot attention and improve object alignment without external supervision.

Let $A^{\text{SA}} \in \mathbb{R}^{L \times N}$ denote the slot attention map from Slot Encoder, where $L$ is the number of visual feature locations (e.g., patch tokens) and $N$ is the number of slots. We also extract the adapter cross-attention probabilities from the UNet, $A^{\text{AD}} \in \mathbb{R}^{L' \times N}$, by collecting the attention weights of the adapter cross-attention layers (averaged over heads). Since $L'$ may differ from $L$, we upsample $A^{\text{AD}}$ to match the resolution of $A^{\text{SA}}$.
We then minimize a binary cross-entropy loss between the two attention maps:
\begin{equation}
\mathcal{L}_{guide} = \mathrm{BCEWithLogits}\!\left(\mathrm{logit}\!\left(\tilde{A}^{\text{AD}}\right),\, A^{\text{SA}}\right), 
\end{equation}
where $\tilde{A}^{\text{AD}}$ denotes the upsampled adapter attention probabilities. 

Following \cite{akan2025slot}, we apply this guidance loss only at a selected UNet block (e.g., the third upsampling block). This guidance encourages consistency between the object partitions discovered by Slot Encoder and the regions attended by the trainable adapter pathway, improving the object-level controllability of the resulting slot representations. Finally, the first training stage optimizes the weighted sum of five losses.
\begin{equation}
\mathcal L_{1}=w_1 \mathcal L_{clip} + w_2 \mathcal L_{diff} + w_3 \mathcal L_{itc} + w_4 \mathcal L_{i2t} + w_5 \mathcal L_{guide}.
\end{equation}




\textbf{Discretization.}
For the second stage, Slot Encoder and Visual Decoder are frozen, and only the Residual Vector Quantizer (RVQ) \cite{lee2022autoregressive} and an additional Transformer block are trained to reconstruct the slot embeddings. Following standard residual vector quantization, the training optimizes an embedding reconstruction loss and a commitment loss.
\begin{equation}
    \mathcal L_{recon}=||S^{\prime}_{emb}-S_{emb}||^2_2,
\end{equation}
\begin{equation}
    \mathcal L_{commit} = \sum^K_{k=1}||Z-sg[\hat Z^{(k)}]||^2_2.
\end{equation}
Here, $S_{emb}$ denotes the original continuous slot embeddings and $S^{\prime}_{emb}$ their reconstruction after RVQ. The RVQ operates over $K$ stages using a shared codebook, where $Z$ and $\hat Z^{(k)}$ represent the pre-quantized and quantized features at depth $k$, respectively, and $\mathrm{sg}[\cdot]$ denotes the stop-gradient operator. We use $K=4$ with codebook size 8192. Detailed information is provided in the supplemental material.

This two-stage design separates object-centric representation learning from discrete codebook learning. By first learning continuous slots with reconstruction, language alignment, and attention guidance objectives, SlotTok obtains semantically structured embeddings before quantization. The second stage then learns to approximate these embeddings with RVQ codes without disturbing the learned slot-object alignment or the pretrained diffusion decoder.

\subsection{Slot-MLLM: Multimodal LLM using Slot Tokens}



Building on SlotTok, we introduce \textbf{Slot-MLLM}, an object-centric generative MLLM that integrates visual and linguistic modalities under a unified autoregressive framework. SlotTok converts an image into $N$ object-centric slots, each quantized by an RVQ with $K$ codebooks. The resulting $N \times K$ discrete indices are flattened into a visual token sequence and added to the LLM vocabulary as a second language. This allows Slot-MLLM to generate visual tokens using the same language modeling head and next-token prediction objective used for text, without modifying the LLM architecture. We use special boundary tokens, \texttt{<boi>} and \texttt{<eoi>}, to indicate the beginning and end of an image-token sequence.

\textbf{Visual Token Representation.}
Slot-MLLM adopts a hybrid visual representation to support both image generation and multimodal understanding. For image generation, we use the discrete slot tokens produced by RVQ, since discrete tokens allow the LLM to autoregressively predict images in the same manner as text. For multimodal understanding, however, directly using discrete visual tokens can introduce quantization errors and discard fine-grained visual information. Therefore, we use the $N$ pre-quantized continuous slot embeddings from SlotTok and project them into the LLM hidden space using a lightweight two-layer MLP projector. This hybrid design allows Slot-MLLM to preserve rich visual features for understanding while maintaining a discrete autoregressive output space for image generation.

\textbf{Multimodal Pre-training.}
We first align SlotTok with the LLM through multimodal pre-training on image--text paired data and interleaved image--text data. Let
$z=(z_1,\ldots,z_T)$ denote a unified multimodal sequence consisting of text tokens and visual slot tokens. Slot-MLLM is trained with a masked next-token prediction objective:
\begin{equation}
\mathcal{L}_{\mathrm{PT}}
=
-\sum_{t=1}^{T}
m_t \log p_{\theta}(z_t \mid z_{<t}),
\label{eq:slot_mllm_pt}
\end{equation}
where $m_t \in \{0,1\}$ is a loss mask that determines whether the token at position $t$ contributes to the training objective. For image--text paired data, we apply the loss only to the target output modality. For example, in text-to-image generation, the loss is applied to the visual slot tokens, whereas in image-to-text generation, the loss is applied to the text tokens. For interleaved image--text data, we apply the loss only to textual tokens, allowing the model to learn multimodal contextual reasoning without forcing unconditional prediction of every image-token segment. During pre-training, SlotTok is kept frozen, and the LLM is adapted using LoRA for efficient multimodal alignment.

\textbf{Supervised Fine-tuning.}
After multimodal pre-training, we perform supervised fine-tuning (SFT) on diverse multimodal instruction-following datasets. The SFT data include image-to-text understanding, text-to-image generation, and image-text-to-image editing examples. Each sample is formatted as a user-assistant conversation, where the user message contains the instruction and optional multimodal context, and the assistant message contains the target response. The SFT objective is defined as
\begin{equation}
\mathcal{L}_{\mathrm{SFT}}
=
-\sum_{t=1}^{T}
\mathbf{1}[z_t \in \mathcal{A}]
\log p_{\theta}(z_t \mid z_{<t}),
\label{eq:slot_mllm_sft}
\end{equation}
where $\mathcal{A}$ denotes the assistant response span. This masking ensures that the model learns to generate the desired response while conditioning on the user instruction and multimodal context, rather than imitating the entire prompt. For understanding tasks, the assistant response consists of text tokens. For generation and editing tasks, the assistant response may contain visual slot tokens enclosed by \texttt{<boi>} and \texttt{<eoi>}. Additional details on dataset composition, prompt templates, and hyperparameter settings are provided in the supplemental material.

%% file: sections/experiments.tex
\subsection{Experimental Setup}

Regarding SlotTok, we adopt DINOv2 \cite{oquab2023dinov2} as our vision encoder due to its proficiency in extracting rich local semantic information. The training dataset for SlotTok includes 70 million image--text pairs. We fix the number of slots to $N=32$, which we found sufficient to cover realistic multi-object scenes while maintaining global semantic coherence and stable text alignment. This choice reflects a trade-off between object coverage and token efficiency, rather than computational speed. Detailed information regarding the datasets and configurations is provided in the supplemental material.


For Slot-MLLM, we adopt Vicuna-7B \cite{zheng2023judging} and Qwen2.5-14B-Instruct \cite{yang2024qwen2} as the base LLMs to demonstrate that SlotTok can be seamlessly applied to any pretrained LLM without architectural modifications. During multimodal pre-training and instruction tuning, we employ Low-Rank Adaptation (LoRA) \cite{hu2021lora} for the LLM while keeping SlotTok frozen. Detailed information regarding datasets, prompt formats, and hyperparameter configurations is provided in the supplemental material.

\begin{table}[t]
\centering
\caption{Image reconstruction performances of visual tokenizers. We compare SlotTok against other diffusion-based visual tokenizers on the COCO Karpathy test set \cite{lin2015microsoft} and ImageNet 50K. Metrics include semantic (CLIP-T), perceptual (DreamSIM, LPIPS), and pixel-level (PSNR, SSIM) similarity. rFID is measured on ImageNet 50K. SlotTok consistently achieves comparable reconstruction quality across most metrics.}
\label{tab:tokenizer}
\resizebox{\columnwidth}{!}{\begin{tabular}{@{}lcccccc@{}}
\toprule
\textbf{Method} & \textbf{CLIP-T ↑} & \textbf{DreamSIM ↓} & \textbf{LPIPS ↓} & \textbf{PSNR ↑} & \textbf{SSIM ↑} & \textbf{rFID ↓} \\ \midrule
LaViT \cite{jin2023unified} & {\bf 25.63} & 0.3185 & 0.6558 & 10.76 & {\bf 0.2910} & {\bf 6.27} \\
SEED \cite{ge2023making}  & 24.45 & 0.4132 & 0.7741 & 9.51  & 0.2568 & 9.99 \\
SlotTok & 25.01 & {\bf 0.2960} & {\bf 0.5327} & {\bf 11.23} & 0.2188 & 9.49 \\ \bottomrule
\end{tabular}}
\end{table}

\subsection{Evaluation of SlotTok}


The image reconstruction experiments assess the effectiveness of SlotTok in preserving visual details when converting images into discrete tokens, as summarized in Table~\ref{tab:tokenizer}. Performance is evaluated through three primary dimensions: semantic similarity to the text description (CLIP-T \cite{radford2021learning}), perceptual similarity (DreamSIM \cite{fu2023dreamsim}, LPIPS \cite{zhang2018unreasonable}), and pixel-level fidelity (PSNR, SSIM \cite{1284395}) between the reconstructed image and the ground-truth image. We compare SlotTok with other widely-used visual tokenizers, including SEED \cite{ge2023making} and LaViT \cite{jin2023unified}, which are trained independently of the LLM and employ diffusion-based decoders. These models closely align with SlotTok's core design and allow for a fair comparison. We further measure rFID \cite{heusel2017gans} on ImageNet 50K to evaluate the distributional realism of reconstructed images. SlotTok achieves competitive rFID, indicating that object-centric slot conditioning preserves image realism.

\begin{figure}[t]
    \centering
    \includegraphics[width=1\linewidth]{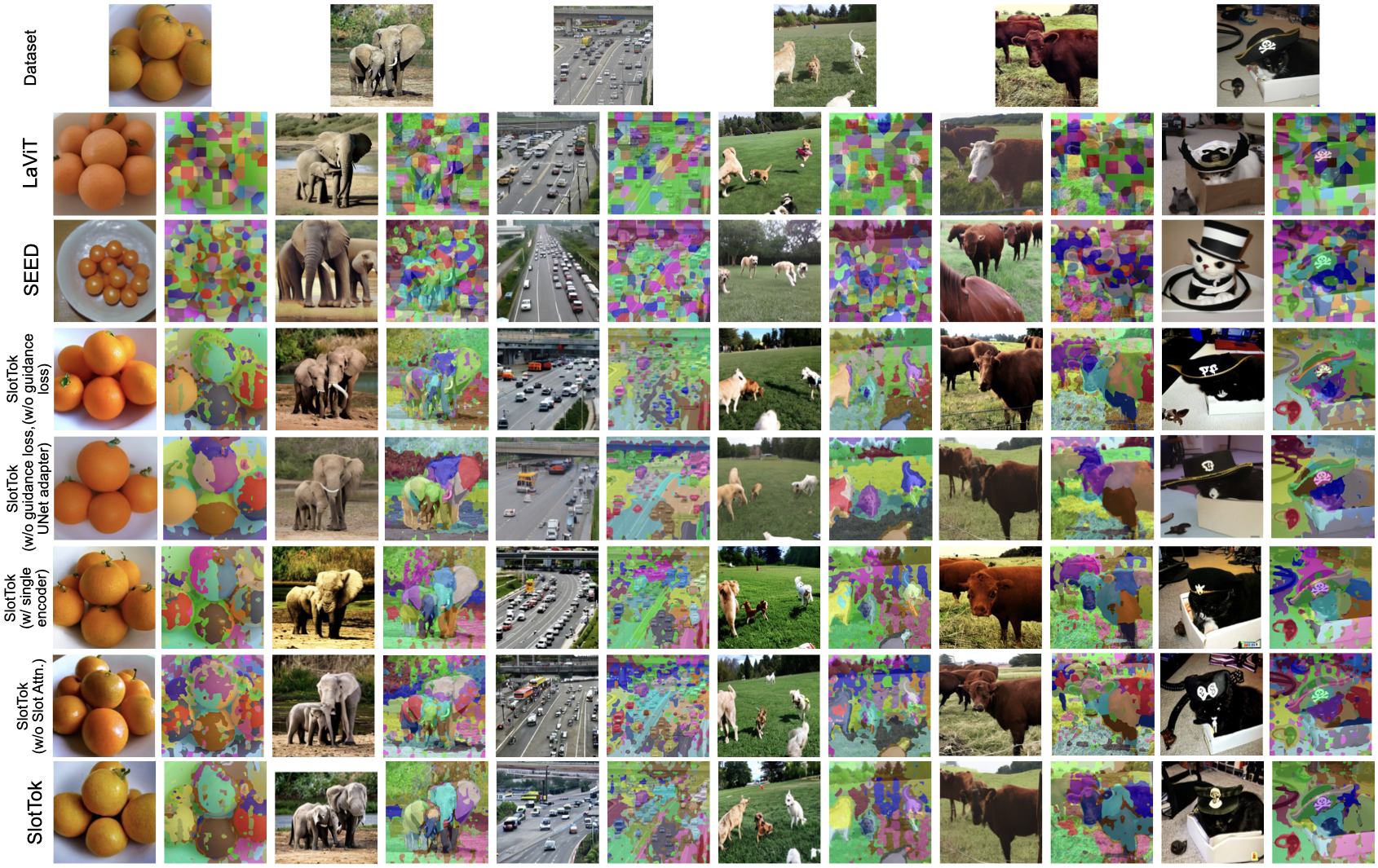}
    \caption{Qualitative results of visual tokenizers. This figure visualizes the reconstructed images generated by various tokenizers, along with the corresponding image regions attended by each token. SlotTok effectively captures object-centric representations by aligning token attention with distinct objects within the image, and can reconstruct images that preserve the original object structures and compositions.
    }
    \label{fig:tokenizer_example}
\end{figure}

\textbf{Qualitative Analysis of Slot Tokens.} Fig.~\ref{fig:tokenizer_example} qualitatively compares reconstructed images and attention maps across different tokenizers. SlotTok preserves original object positions and overall scene composition in reconstruction, while its attention maps clearly concentrate on individual objects, demonstrating effective object-centric token formation. In contrast, despite employing a larger number of visual tokens, LaViT fails to maintain consistent spatial layouts, often producing reconstructions with shifted object positions. SEED mainly captures global scene-level semantics, with attention maps broadly distributed over the image rather than localized to distinct objects. Similarly, replacing slot attention with standard cross-attention in SlotTok results in dispersed attention patterns and degraded object alignment. These observations underscore the critical role of slot attention in enabling object-centric visual tokenization.

\subsection{Analysis of SlotTok}

To verify whether each visual token generated by SlotTok effectively encodes distinct and detailed image information, we conduct a randomized replacement analysis. The core premise is that if each token distinctly represents specific regions or objects, such removal would lead to a significant loss of information and a substantial drop in performance. Conversely, if visual information is overlapping or redundant between tokens, the remaining representations will preserve sufficient context, resulting in only minor performance changes.
In this setup, we randomly select 50\% of the 32 visual representations. For the reconstruction task, we replace the selected discrete visual tokens with random indices sampled from the codebook (indices 0-8191) and evaluate the reconstruction performance based on these altered tokens. This approach ensures that the perturbed information cannot be utilized during processing, allowing us to quantify the necessity of each individual slot.

\begin{table}[t]
\centering
\caption{Impact of randomly replacing visual tokens on image reconstruction across SlotTok variants. Performance metrics after token replacement are indicated with the prefix ‘Drop’ (e.g., Drop LPIPS), and the values in parentheses denote the changes in performance.}
\label{tab:token_drop_test}
\resizebox{\columnwidth}{!}{
\begin{tabular}{@{}lcccc@{}}
\toprule
\textbf{Method}    & \textbf{LPIPS ↓} & \textbf{Drop LPIPS ↑} & \textbf{PSNR ↑} & \textbf{Drop PSNR ↓} \\ \midrule
SlotTok            & 0.5327           & \textbf{0.6905 (+ 29.6\%)}   & 11.23           & \textbf{8.78 (-21.8\%)}     \\
w/o Slot Attention & \textbf{0.5306}           & 0.6789 (+27.9\%)    & \textbf{11.41}           & 9.19 (-19.5\%)     \\ \bottomrule
\end{tabular}
}
\end{table}

As shown in Table~\ref{tab:token_drop_test}, visual tokens generated by SlotTok exhibit a larger performance drop in reconstruction performance upon replacement compared to the SlotTok without slot attention. These results indicate that the proposed slot attention successfully partitions the visual scene into distinct and non-redundant visual tokens, where each slot encodes specific visual details. 

Fig.~\ref{fig:tokenizer_example} further visualizes how different tokenizer variants affect slot attention patterns. Variants without slot attention or attention guidance produce more dispersed and less object-aligned attention maps, whereas the full SlotTok model yields more coherent object-centric regions during reconstruction.

\begin{table*}[t!]
\centering
\caption{Evaluation on multimodal understanding benchmarks comparing Slot-MLLM against existing MLLMs. * denotes performance obtained by our own evaluation using publicly available checkpoints. $\dagger$ denotes scores where, due to the model’s lack of support for multi-image inputs, performance on multi-image samples is measured by randomized responses. Models are categorized by their pre-training data scale, and the best performance within each group among 7B-scale models is highlighted in bold.}
\label{tab:understanding}
\resizebox{\textwidth}{!}{%
\begin{tabular}{@{}lcccccccccc@{}}
\toprule
\textbf{Method} &
  \textbf{LLM} &
  \textbf{PT Data Size} &
  \textbf{GQA↑} &
  \textbf{MME-P↑} &
  \textbf{POPE↑} &
  \textbf{MMB↑} &
  \textbf{MMMU↑} &
  \textbf{SEED-IMG↑} &
  \textbf{NaturalBench↑} &
  \textbf{SEEDBench-2-Plus↑} \\ \midrule
LaVIT &
  LLaMA-7B &
  100M &
  47.8* &
  964.5* &
  \textbf{83.4*} &
  34.2* &
  24.6 $\dagger$ &
  35* &
  52.5* &
  27.5* \\
SEED-LLaMA &
  Vicuna-7B &
  77M &
  52.4* &
  1123.9* &
  78* &
  45.5* &
  26.8* &
  48.6* &
  \textbf{57.5*} &
  24.2* \\
UniTok &
  LLaMA-2-7B &
  70M &
  \textbf{61.1*} &
  \textbf{1465.7*} &
  83.2* &
  \textbf{74.6*} &
  \textbf{33.2 $\dagger$} &
  \textbf{67.6*} &
  50.6* &
  \textbf{38.6*} \\ \midrule
MorphTokens &
  Vicuna-7B &
  30M &
  56.8 &
  \textbf{1477.7} &
  - &
  - &
  - &
  - &
  - &
  - \\
VILA-U &
  LLaMA-2-7B &
  23M &
  \textbf{58.3*} &
  1336.2 &
  82.7* &
  \textbf{73.6*} &
   29.2* &
  \textbf{61.9*} &
   63.1* &
  \textbf{32.4*} \\
LWM &
  LLaMA-2-7B &
  - &
  44.8 &
  - &
  75.2 &
  - &
  - &
  - &
  - &
  - \\
Slot-MLLM (7B) &
  Vicuna-7B &
  25M &
  57.3 &
  1225.2 &
  \textbf{83.0} &
  66.8 &
  \textbf{33.2} &
  58.7 &
  \textbf{66.1} &
  31.2 \\ \midrule
Slot-MLLM (14B) &
  Qwen2.5-14B-Instruct &
  25M &
  58.8 &
 1468.3 &
  82.2 &
   74.9 &
   46.7 &
   62.9 &
   69.8 &
   34.4 \\ \bottomrule
\end{tabular}
}
\end{table*}

\subsection{Evaluation of Slot-MLLM}
This section evaluates Slot-MLLM against unified MLLMs across both image understanding and generation benchmarks. We focus on integrated models that perform both tasks within a single architecture, excluding models like TokenFlow \cite{qu2025tokenflow} that utilize separate checkpoints specialized for each task. LaViT~\cite{jin2023unified} and SEED-LLaMA~\cite{ge2023making} serve as our primary baselines, as they are the most closely related unified MLLMs that also employ diffusion-based visual decoders.

\textbf{Multimodal Understanding.} Table~\ref{tab:understanding} shows the comparison between Slot-MLLM and existing unified MLLMs across various zero-shot multimodal understanding benchmarks. These evaluations, including GQA \cite{hudson2019gqa}, MME \cite{fu2023mme}, POPE \cite{li2023evaluating}, MMBench \cite{liu2024mmbench}, MMMU \cite{yue2024mmmu}, SEED-Bench-IMG \cite{li2024seed}, NaturalBench \cite{li2024naturalbench}, and SEEDBench-2-Plus \cite{li2024seed2plus}, assess capabilities ranging from basic object perception to high-level logical reasoning. 

To ensure a fair comparison, we categorize models into two groups based on their pre-training (PT) data scale, using a threshold of 50M samples. Experimental results indicate that Slot-MLLM outperforms LaViT and SEED-LLaMA on most benchmarks. Notably, our model achieves a significant score of 66.1 on NaturalBench, surpassing both UniTok \cite{ma2025unitok} and VILA-U \cite{wu2024vila}. This success is largely attributed to slot attention, which effectively segments images into object-centric visual tokens to improve the ability of the LLM to comprehend visual scenes, especially in tasks requiring understanding of object recognition, attribute binding, and logical reasoning. 

While UniTok and VILA-U report higher scores on certain benchmarks, these gaps are primarily driven by architectural design and training scale. VILA-U and UniTok utilize VQ-based tokenizers that generate significantly more tokens per image--typically 256 or more--compared to the 32 tokens produced by SlotTok. Furthermore, VILA-U requires a large-scale dataset comprising 700M image--text pairs for tokenizer training, which is about 10 times larger than ours.

\textbf{Multimodal Generation.} Table~\ref{tab:generation} compares Slot-MLLM to unified MLLMs on text-to-image generation and image editing benchmarks. T2I-CompBench \cite{huang2023t2i}, GenEval \cite{ghosh2023geneval}, and GenAI-Bench \cite{li2024genai} evaluate the accuracy and compositional fidelity of generated images described in textual prompts.
MagicBrush \cite{zhang2023magicbrush} assesses the model’s capability for localized image editing using metrics such as CLIP similarity (CLIP-I) and pixel-level differences (L1) between generated images and ground-truth images. Additionally, GIE-Bench \cite{qian2025gie} provides a grounded evaluation of image editing by measuring functional correctness via multiple-choice questions and content preservation using CLIP similarity in background regions outside the object-aware mask to ensure scene integrity. 

\begin{figure}[t!]
    \centering
    \includegraphics[width=1\linewidth]{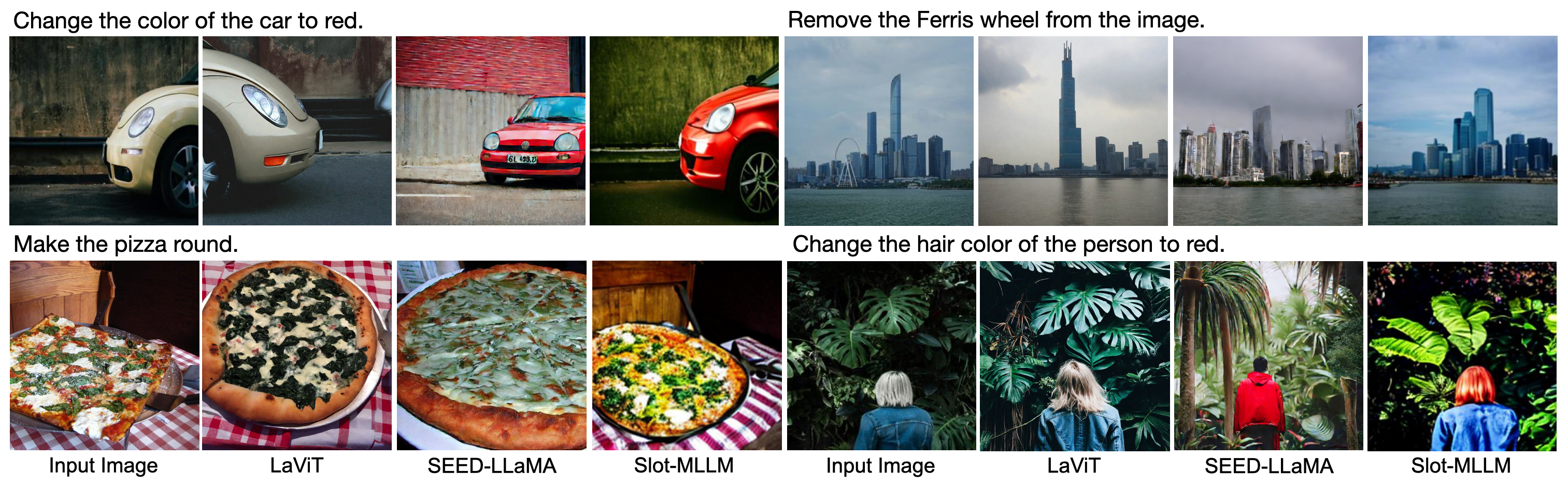}
    \caption{Qualitative results on image editing tasks. Slot-MLLM effectively modifies specific objects described by text prompts while preserving the overall image composition and context.}
    \label{fig:edit-sample}
\end{figure}

Notably, Slot-MLLM consistently outperforms SEED-LLaMA and LaVIT across the majority of generation and editing tasks. This is particularly noteworthy as both SEED-LLaMA and LaViT utilize diffusion-based decoders similar to our approach, yet were trained on larger scales. Within the smaller-scale group, our model achieves a leading score on GenEval, demonstrating that object-centric visual tokens successfully capture detailed semantics such as color, position, quantity, and texture. Qualitative examples in Fig.~\ref{fig:edit-sample} further illustrate that Slot-MLLM effectively maintains background consistency and handles object-level manipulations that others fail to demonstrate. Furthermore, scaling the backbone LLM consistently improves both understanding and generation performance, demonstrating that Slot-MLLM benefits from stronger LLMs while preserving the efficiency of its slot-based architecture. More generated results are provided in the supplemental material.

Models using discrete tokens like LaViT, VILA-U, and UniTok struggle with editing tasks. Although these models exhibit strong general text-to-image generation by employing a larger number of visual tokens, LaViT falls short compared to Slot-MLLM in editing tasks. VILA-U and UniTok lack supervised fine-tuning specifically tailored for editing, resulting in poor performance in localized editing scenarios. We do not include reproduced results for MorphTokens because its code and model checkpoints are not publicly available. These results underscore the practical advantage of SlotTok, which achieves a superior balance between generative quality and computational efficiency.



\begin{table*}[t]
\centering
\caption{Evaluation on text-to-image and editing benchmarks. As VILA-U and UniTok are not trained on multimodal prompts for image editing (IT2I), they are excluded from the editing benchmarks.}
\label{tab:generation}
\resizebox{\textwidth}{!}{%
\begin{tabular}{@{}lccccccccccccc@{}}
\toprule
\multirow{2}{*}{\textbf{Method}} &
  \multirow{2}{*}{\textbf{LLM}} &
  \multirow{2}{*}{\textbf{PT Data Size}} &
  \multirow{2}{*}{\textbf{GenEval↑}} &
  \multicolumn{3}{c}{\textbf{T2I-CompBench↑}} &
  \multicolumn{2}{c}{\textbf{GenAI-Bench↑}} &
  \multicolumn{2}{c}{\textbf{MagicBrush}} &
  \multicolumn{2}{c}{\textbf{GIEBench}} \\ \cmidrule(l){5-7} \cmidrule(lr){8-9} \cmidrule(lr){10-11} \cmidrule(lr){12-13}
 &
   &
   &
   &
  color &
  shape &
  texture &
  basic &
  advanced &
  CLIP-I ↑ &
  L1 ↓ &
  MCQA ↑ &
  CLIP ↑ \\ \midrule
LaVIT &
  LLaMA-7B &
  100M &
  0.48* &
  0.2730* &
  0.3191* &
  0.3988* &
  0.78* &
  0.62* &
  \textbf{78.5*} &
  \textbf{28.6*} &
  20.57* &
  \textbf{91.5*} \\
SEED-LLaMA &
  Vicuna-7B &
  77M &
  0.29* &
  0.1741* &
  0.2033* &
  0.2361* &
  0.52* &
  0.49* &
  77.9* &
  28.8* &
  \textbf{24.00*} &
  88.6* \\
UniTok &
  LLaMA-2-7B &
  70M &
  \textbf{0.59*} &
  \textbf{0.7707*} &
  \textbf{0.5298*} &
  \textbf{0.6521*} &
  \textbf{0.82*} &
  \textbf{0.66*} &
  - &
  - &
  - &
  - \\ \midrule
MorphTokens &
  Vicuna-7B &
  30M &
  - &
  - &
  - &
  - &
  - &
  - &
  \textbf{87.9} &
  \textbf{7.6} &
  - &
  - \\
VILA-U &
  LLaMA-2-7B &
  23M &
  0.42* &
  \textbf{0.5389*} &
  \textbf{0.4123*} &
  \textbf{0.5079*} &
  \textbf{0.76*} &
  \textbf{0.64*} &
  - &
  - &
  - &
  - \\
LWM &
  LLaMA-2-7B &
  - &
  0.47 &
  - &
  - &
  - &
  - &
  - &
  - &
  - &
  - &
  - \\
Slot-MLLM (7B) &
  Vicuna-7B &
  25M &
  \textbf{0.53} &
  0.5308 &
  0.3589 &
  0.4616 &
  0.67 &
  0.59 &
  81.8 &
  22.8 &
  \textbf{23.9} &
  \textbf{94.2} \\ \midrule
Slot-MLLM (14B) &
  Qwen2.5-14B-Instruct &
  25M &
  0.57 &
  0.5885 &
  0.4025 &
  0.5274 &
  0.58 &
  0.54 &
   81.7 &
   22.8 &
   20.2&
   94.2 \\ \bottomrule
\end{tabular}
}
\end{table*}

\begin{table*}[t]
\centering
\caption{Evaluation on multimodal understanding benchmarks comparing Slot-MLLM against understanding-only models.}
\label{tab:understanding_specialist}
\resizebox{\textwidth}{!}{%
\begin{tabular}{@{}lccccccccc@{}}
\toprule
\textbf{Method}          & \textbf{LLM}                  & \textbf{GQA↑}      & \textbf{MME-P↑}       & \textbf{POPE↑} & \textbf{MMB↑}  & \textbf{MMMU↑}       & \textbf{SEED-IMG↑} & \textbf{NaturalBench↑} & \textbf{SEEDBench-2-Plus↑} \\ \midrule
InstructBLIP \cite{Dai2023InstructBLIPTG}    & Vicuna-7B            & 49.2      & -            & -     & -     & -            & -         & 59.1             & -                 \\
LLaVA-NeXT \cite{liu2024llavanext}      & Vicuna-7B            & \textbf{64.2}      & 1519.0         & 86.5  & 67.4  & 35.8         & 70.2      & 70.2             & -                 \\
ShareGPT4V \cite{chen2024sharegpt4v}     & Vicuna-7B            & -         & 1567.4       & -     & 68.8  & -            & 69.7      & 68.4             & -                 \\
LLaVA-OneVision \cite{li2024llavaone}  & Qwen2-7B             & -         & \textbf{1580.0}         & -     & 80.8  & 48.8         & \textbf{75.4}      & \textbf{77.2}             & -                 \\ 
Qwen2.5-VL      & Qwen2.5-7B           & -         & -   & -     & 83.5  & \textbf{58.6}         & -         & -             & \textbf{70.4}              \\
InternVL2.5 \cite{chen2024expanding}    & InternLM2.5-7B       & -         & - & \textbf{90.6}  & \textbf{84.6}  & 56.0         & -         & -             & 69.7              \\ \midrule
Slot-MLLM (7B)       & Vicuna-7B            & 57.3      & 1225.2       & 83.0  & 66.8  & 33.2         & 58.7      & 66.1          & 31.2              \\
Slot-MLLM (14B)       & Qwen2.5-14B-Instruct & 58.8      & 1468.3       & 82.2  & 74.9  & 46.7         & 62.9      & 69.8          & 34.4               \\ \bottomrule
\end{tabular}%
}
\end{table*}

\begin{table}[t]
\centering
\caption{Evaluation on text-to-image benchmarks comparing Slot-MLLM against generation-only models.}
\label{tab:generation_specialist}
\resizebox{\columnwidth}{!}{%
\begin{tabular}{@{}lccccc@{}}
\toprule
\multirow{2}{*}{\textbf{Method}} & \multirow{2}{*}{\textbf{Type}} & \multirow{2}{*}{\textbf{GenEval↑}} & \multicolumn{3}{c}{\textbf{T2I-CompBench↑}} \\ \cmidrule(l){4-6} 
          &                &      & color  & shape  & texture \\ \midrule
SD v2.1   & Diffusion      & 0.50 & 0.5694 & 0.4495 & 0.4982  \\
SD-XL     & Diffusion      & 0.55 & 0.5879 & \textbf{0.4687} & \textbf{0.5299}  \\ \midrule
Slot-MLLM (7B) & LLM-based      & 0.53 & 0.5308 & 0.3589 & 0.4616  \\
Slot-MLLM (14B) & LLM-based      & \textbf{0.57} & \textbf{0.5885} & 0.4025 & 0.5274  \\ \bottomrule
\end{tabular}
}
\end{table}

\textbf{Comparison with specialist models.} Tables~\ref{tab:understanding_specialist} and ~\ref{tab:generation_specialist} compare Slot-MLLM with task-specific understanding-only and generation-only models. Although specialist models such as SD-XL, Qwen2.5-VL, and InternVL2.5 achieve stronger results on their respective tasks, they are trained at substantially larger scales and optimized for a single objective. In contrast, Slot-MLLM supports both visual understanding and image generation within a unified framework using compact object-centric visual tokens.


\subsection{Ablation Study} \label{ablation-study}

\begin{table}[t!]
\centering
\caption{Tokenizer ablation on reconstruction, distributional realism, and object-centric metrics. ARI and mIoU are computed from slot attention maps on COCO 2017 panoptic validation images.}
\label{tab:tokenizer_ablation}
\resizebox{\columnwidth}{!}{%
\begin{tabular}{@{}lcccccc@{}}
\toprule
\textbf{Method} & \textbf{CLIP-T ↑} & \textbf{DreamSIM ↓} & \textbf{PSNR ↑} & \textbf{rFID ↓} & \textbf{ARI ↑} & \textbf{mIoU ↑} \\ \midrule
SlotTok                         & 25.01          & 0.2960          & 11.23          & 9.49          & \textbf{0.2028} & \textbf{0.2594} \\
w/o guidance loss               & 25.23          & \textbf{0.2880} & 11.70          & \textbf{8.74} & 0.1899          & 0.2495          \\
w/o guidance loss and UNet adapter & 24.95       & 0.3166          & \textbf{12.14} & 10.89         & 0.2018          & 0.2516          \\
w/ single encoder               & \textbf{25.38} & 0.2981          & 11.34          & 8.78          & 0.1779          & 0.2394          \\
w/o Slot Attention              & 24.91          & 0.3050          & 11.41          & 9.74          & 0.1381          & 0.2108          \\ \bottomrule
\end{tabular}%
}
\end{table}

\begin{table}[t!]
\centering
\caption{Ablation results on SlotTok and Slot-MLLM. We report tokenizer reconstruction, multimodal understanding, and generation performance under different design choices. VQGAN-MLLM is excluded from the generation evaluation due to image generation failure.}
\label{tab:ablation_mllm}
\resizebox{\columnwidth}{!}{%
\begin{tabular}{@{}lcccccc@{}}
\toprule
\multirow{2}{*}{\textbf{Method}} & \multicolumn{2}{c}{\textbf{Tokenizer}} & \multicolumn{2}{c}{\textbf{Understanding}} & \multicolumn{2}{c}{\textbf{Generation}} \\ \cmidrule(l){2-3} \cmidrule(lr){4-5} \cmidrule(lr){6-7} 
                          & DreamSIM ↓ & PSNR ↑         & GQA↑      & POPE ↑ & GenEval ↑ & MagicBrush ↑ \\ \midrule
Slot-MLLM                 & 0.2960              & 11.23          & \textbf{57.3}      & \textbf{83.0}   & \textbf{0.53}     & \textbf{81.8}                 \\
w/o guidance loss &  \textbf{0.2880}             & \textbf{11.70}          &  57.1    &  \textbf{83.0}   &  0.50    &  \textbf{81.8}                \\
w/ single encoder         & 0.2981              & 11.34          &  56.7    &  79.9   & 0.49      & 81.4                 \\
w/o Slot Attention        & 0.3050              & 11.41          & 55.3      & 79.3   & 0.48      & 81.4                 \\
w/ Discrete Inputs (Und.) & 0.2960              & 11.23          & 54.5      & 79.4   & 0.52      & \textbf{81.8}                 \\ \midrule
VQGAN-MLLM                & \textbf{0.1855}     & \textbf{13.80} & 36.1 & 67.3 &   -      &        -            \\ \bottomrule
\end{tabular}
}
\end{table}

We conduct ablation studies to analyze the key design choices underlying SlotTok and Slot-MLLM, focusing on three primary questions:
(1) which tokenizer components are critical for effective object-centric tokenization,
(2) whether the inductive bias of slot attention provides a meaningful advantage over standard cross-attention mechanisms, and (3) the optimal strategy for integrating visual representations into MLLMs.



\textbf{Which Components Matter for Object-Centric Tokenization?} Table~\ref{tab:tokenizer_ablation} analyzes the architectural elements necessary for effective object-centric tokenization. We evaluate the quality of object-level information using Adjusted Rand Index (ARI)~\cite{hubert1985comparing} and mean Intersection over Union (mIoU)~\cite{everingham2010pascal} on the COCO 2017 panoptic validation set. Specifically, ARI captures the consistency of slot assignments within ground-truth (GT) objects, while mIoU measures the spatial alignment between slots and GT object segments after unsupervised matching. These metrics are computed from attention maps extracted from the second-to-last slot attention layer of the Slot Encoder, ensuring a stable representation prior to image decoding.

As shown in Table~\ref{tab:tokenizer_ablation}, utilizing standard cross-attention (Row 5) yields comparable reconstruction quality in terms of PSNR but suffers from significantly lower ARI and mIoU, indicating a failure to form distinct object-level groupings. While using a single DINOv2 encoder (Row 4)--instead of the dual DINOv2 and CLIP setup--improves distributional realism as measured by rFID, it still lacks the object-level consistency. Furthermore, training without guidance loss while directly fine-tuning the UNet (Row 3) achieves the highest pixel-level fidelity but substantially degrades rFID, ARI, and mIoU. We also observe that adding lightweight UNet adapters without guidance loss (Row 2) helps restore rFID, yet it nonetheless weakens object alignment. This suggests that without proper guidance, the pretrained diffusion prior is partially overwritten, sacrificing global realism. Ultimately, combining adapter-based conditioning with attention guidance loss (Row 1) achieves the optimal balance, preserving reconstruction fidelity while producing the most structurally consistent representations.


Although the absolute ARI/mIoU scores are modest, we use them only as diagnostic indicators under unconstrained natural images, not as the primary objective of SlotTok. More importantly, Table~\ref{tab:ablation_mllm} shows that removing the guidance loss improves reconstruction metrics such as PSNR but degrades GQA and GenEval, suggesting that object-level alignment provides benefits beyond pixel-level fidelity.

\textbf{How Many Slots Are Needed for Object-Centric Tokenization?}
Table~\ref{tab:slot_token_num_ablation} analyzes the effect of the number of slots in SlotTok. We vary the number of slots from 16 to 64 and evaluate both reconstruction-oriented metrics and object-centric metrics. Increasing the number of slots generally improves reconstruction quality, as reflected by better CLIP-T, DreamSIM, LPIPS, PSNR, and SSIM scores. This is expected because larger slots provide higher representational capacity and allow the tokenizer to preserve finer visual details.

However, object-centric metrics exhibit a different trend. Increasing the number of slots from 16 to 32 improves both ARI and mIoU, but further increasing the number of slots to 64 substantially degrades them. This suggests that excessive slots tend to over-partition coherent objects into smaller sub-regions rather than preserving object-level units. Therefore, a larger number of slots is not always preferable for object-centric tokenization, even when it improves reconstruction fidelity. Based on this trade-off, we use 32 slots as the default configuration, which provides the best object-centric alignment among the tested settings while maintaining competitive reconstruction quality and token efficiency.



\begin{table}[t!]
\centering
\caption{Effect of the number of slots in SlotTok. We compare reconstruction-oriented and object-centric metrics under different slot budgets.}
\label{tab:slot_token_num_ablation}
\resizebox{\columnwidth}{!}{%
\begin{tabular}{@{}lccccccc@{}}
\toprule
\textbf{Method} & \textbf{CLIP-T↑} & \textbf{DreamSIM↓} & \textbf{LPIPS↓} & \textbf{PSNR↑} & \textbf{SSIM↑}  & \textbf{ARI↑} & \textbf{mIoU↑} \\ \midrule
SlotTok      & 26.12 & 0.2264 & 0.4538 & 12.71 & 0.2651 & \textbf{0.2028} & \textbf{0.2594} \\
w/ 16 slots & 25.36 & 0.2669 & 0.4818 & 12.17 & 0.2628 & 0.1909          & 0.2235          \\
w/ 64 slots & \textbf{26.40}   & \textbf{0.2103}    & \textbf{0.3844} & \textbf{13.45} & \textbf{0.3021} & 0.1066         & 0.2035          \\ \bottomrule
\end{tabular}%
}
\end{table}

\begin{table}[t]
\centering
\caption{Targeted evaluation on spatially sensitive understanding and generation tasks. GenEval-Position denotes the position subcategory of GenEval. GQA-Spatial is a small subset of 316 GQA validation questions filtered for spatial relationship queries.
}

\label{tab:spatial_ablation}
\begin{tabular}{lcc}
\toprule
Method & GenEval-Position & GQA-Spatial \\
\midrule
Slot-MLLM & \textbf{45.8} & \textbf{75.3} \\
w/o Slot Attention & 31.5 & 66.5 \\
w/o guidance loss & 40.5 & 68.8 \\
\bottomrule
\end{tabular}
\end{table}

\textbf{Is Slot Attention Effective?} We next examine whether object-centric tokenization is essential for downstream multimodal learning by comparing SlotTok with alternative tokenizers that lack this specific architectural constraint. We contrast slot attention with standard cross-attention, and further compare our approach against a VQGAN-based tokenizer to evaluate the impact of high-fidelity reconstruction versus object-level information.

As demonstrated in Table~\ref{tab:ablation_mllm}, weak object-centric information in standard cross-attention mechanisms (Row 4) directly leads to a significant performance drop in both understanding and generation tasks that require compositional reasoning and object-level control. This disparity is even more pronounced when compared with the VQGAN-based tokenizer. Although VQGAN achieves superior pixel-level reconstruction, its lack of semantic information results in significantly worse performance across downstream benchmarks (Row 6). These results confirm that high reconstruction fidelity does not inherently guarantee semantic understanding; for tasks requiring reasoning, visual tokens must provide an object-aware representation that aligns with linguistic concepts. Consequently, our findings establish that our object-centric tokenization serves as a critical intermediate representation for MLLMs, rather than being a mere byproduct of improved image synthesis.


\textbf{The Effect of Discretizing Slot Embeddings.}
Finally, we analyze the impact of using pre-quantization continuous slot embeddings versus discrete slot tokens for multimodal understanding within our unified autoregressive framework. Our results in Table~\ref{tab:ablation_mllm} show that continuous embeddings (Row 1) consistently yield superior performance in understanding benchmarks compared to using their discrete counterparts (Row 5). This performance gap confirms that the discretization process introduces quantization errors. However, discrete tokens remain indispensable for the generative output space, as they enable the LLM to perform autoregressive next-token prediction for image synthesis. By employing this hybrid design, Slot-MLLM utilizes the full richness of visual features for complex reasoning while maintaining the quality of compositional image generation.


\textbf{Does SlotTok Preserve Spatial Relationships?}
We conduct targeted evaluations on spatially sensitive tasks using the position subcategory of GenEval and a small GQA-Spatial subset of 316 validation questions filtered for spatial relationship queries. As shown in Table~\ref{tab:spatial_ablation}, Slot-MLLM consistently outperforms variants without slot attention and without guidance loss on both GenEval-Position and GQA-Spatial, suggesting that object-centric slot tokens preserve spatial information useful for relative object reasoning. These results indicate that object-centric tokenization provides a structured visual interface for relational reasoning, beyond simply preserving low-level reconstruction fidelity.

These results also clarify why object-centric tokenization should be evaluated beyond reconstruction metrics alone. While reconstruction quality measures how faithfully visual content is preserved, downstream MLLM tasks require tokens that expose object identity, attributes, and relations in a form accessible to the language model. SlotTok addresses this requirement by combining compact slot-based grouping with language-aligned and generation-compatible visual representations. Overall, these ablations demonstrate that effective object-centric visual tokenization for MLLMs requires coordinated design choices across slot formation, tokenizer training, and representation integration, rather than isolated improvements in reconstruction quality.

%% file: sections/conclusion.tex
We investigate how object-centric slot attention can be extended to multimodal large language models through SlotTok. We show that directly applying tokenization optimized for image reconstruction is insufficient for downstream MLLM tasks, and that adapting slot attention to better align with language supervision and strengthening object-centric representations is necessary. By modifying slot-based tokenization to support both continuous and discrete representations, SlotTok enables stable integration of object-centric visual tokens into generative MLLMs. Experimental results demonstrate that this extension of slot attention leads to competitive performance in multimodal understanding and generation tasks, highlighting the potential of object-centric representations for MLLMs.

%% file: sections/acknowledgments.tex
This work was partly supported by Institute of Information \& communications Technology Planning \& Evaluation (IITP) grant funded by the Korea government(MSIT) (No. RS-2019-II190079, Artificial Intelligence Graduate School Program(Korea University), 50\%), and the National Research Foundation of Korea(NRF) grant funded by the Korea government(MSIT) (No. RS-2024-00353007, 50\%).